\documentclass[runningheads]{llncs}
\usepackage[utf8]{inputenc} 
\usepackage[T1]{fontenc}    
\usepackage{hyperref}       
\usepackage{url}            
\usepackage{booktabs}       
\usepackage{amsfonts}       
\usepackage{nicefrac}       
\usepackage{microtype}      
\usepackage{graphicx}
\usepackage{authblk}

\begin{document}
\title{Predicting Auction Price of Vehicle License Plate with Deep Residual Learning}
\titlerunning{Predicting Price of License Plate}
%
\author{Vinci Chow\inst{1}\orcidID{0000-0003-0754-5348}}
\authorrunning{V. Chow}
%
\institute{The Chinese University of Hong Kong, Shatin, Hong Kong \\
\email{vincichow@cuhk.edu.hk}}
\maketitle              
\begin{abstract}
Due to superstition,
license plates with desirable combinations of characters are highly sought after in China, 
fetching prices that can reach into the millions in government-held auctions. 
Despite the high stakes involved, there has been essentially no attempt to provide price estimates 
for license plates.
We present an end-to-end neural network model that simultaneously predict the auction price, 
gives the distribution of prices and produces latent feature vectors.
While both types of neural network architectures we consider outperform simpler machine learning methods,
convolutional networks outperform recurrent networks for comparable training time or model complexity. 
The resulting model powers our online price estimator and search engine.

\end{abstract}

\section{Introduction}

Chinese society place great importance on numerological superstition. 
Numbers such as 2 and 8 are often used solely because of the desirable qualities they represent 
(easiness and prosperity, respectively). 
For example, the Beijing Olympic opening ceremony occurred on 2008/8/8 at 8 p.m., 
while the Bank of China (Hong Kong) opened for business on 1988/8/8. 
Because license plates represent one of the most public displays of numbers for many people, 
people are willing an enormous amount of money for license plates with desirable combinations of characters. 
Local governments often auction off such license plates to generate public revenue. 
The five most expensive plates ever auctioned in Hong Kong have each sold for over US\$1 million.

Unlike the auctioning of other valuable items, license plates generally do not come with a price estimate,
even though price estimates have been shown to be a significant factor affecting the sale price 
\cite{10.1257/jep.3.3.23,10.2307/1911865}.
It is also very difficult to discover which plates are available at what price because auction outcomes are not 
always available online. 
In Hong Kong, the government only provides the results of the three most recent auctions online, despite
having hosted monthly auctions for several decades. 

We build an online service, \textit{markprice.ai}, that seeks to fill this gap by providing 
three specific features:


\begin{enumerate}
\item
\textit{Price estimation.} 
The primary service we provide is price estimation for license plate auctions in Hong Kong. 
In Hong Kong, license plates consist of either a two-letter prefix or no prefix, 
followed by up to four digits (e.g., HK 1, BC 6554, or 138).
Plates can be desirable because the characters rhythm with auspicious Chinese phrases (
e.g. ``168'' rhythms with ``all the way to prosperity'', while ``186'' does not  rhythm with anything)
or have visual appeal (e.g. ``2112'' is symmetric, while ``2113'' is not.) 
The model learns which features are valuable from the training data, allowing it to then
generate a predicted price for the specific combination of characters a user enters.

\item
\textit{Distribution of predicted prices.}
Even with a perfect model, there will be variation in the realized price due to factors we cannot capture.
We do not wish to give the site's users a false impression of certainty, so it is important that we provide
a distribution of possible outcomes. The main challenge here is providing a distribution for extremely expensive plates, 
for which there are very few examples to use to generate a distribution from.

\item
\textit{Search engine.}
Beyond the price estimation service, we also want to help users research and discover plates that
they might be interested in. We provide a way for users to not only search for past records of a
specific plate, but also to discover other plates that are reasonably similar.
\end{enumerate}

We detail our solution to the above three targets in this study. 
Our model is powered by a neural network that generates a distribution of predicted prices for a given license plate.
In the process, the network also learns to generate latent feature vectors, which we extract to construct
our search engine. 
With both linguistic and visual factors affecting a plate's value, 
our primary focus is on the relative performance of recurrent networks versus convolutional networks
for our task.
Our results suggest that while both neural network architecture outperform simpler machine learning methods,
convolutional networks outperform recurrent networks for comparable training time or model complexity. 

\section{Related studies}
\label{sec-related_studies}

Early studies into the prices of license plates use hedonic regressions with a larger number of handpicked features 
\cite{Woo1994389,Woo200835,Ng2010293}. 
Due to their reliance on ad-hoc features, these models adapt poorly to new data, 
such as when plates with new combinations of characters are auctioned off for the first time. 
In contrast, our model is able to learn the value of license plates from their prices. 
As we demonstrate below, no handpicked feature is needed to achieve high prediction accuracy, 
although the presence of such features does improve the consistency of the learned features.
\cite{license-plate-RNN} attempts to model prices with a standard character-level recurrent neural network,
achieving higher accuracy than that previous studies. 
In this paper we utilizes more advanced neural network designs, resulting in significant improvement in 
accuracy and consistency.

\section{Model}

\subsection{Overall structure}

\begin{figure}
\centering
\includegraphics[width=4.5in]{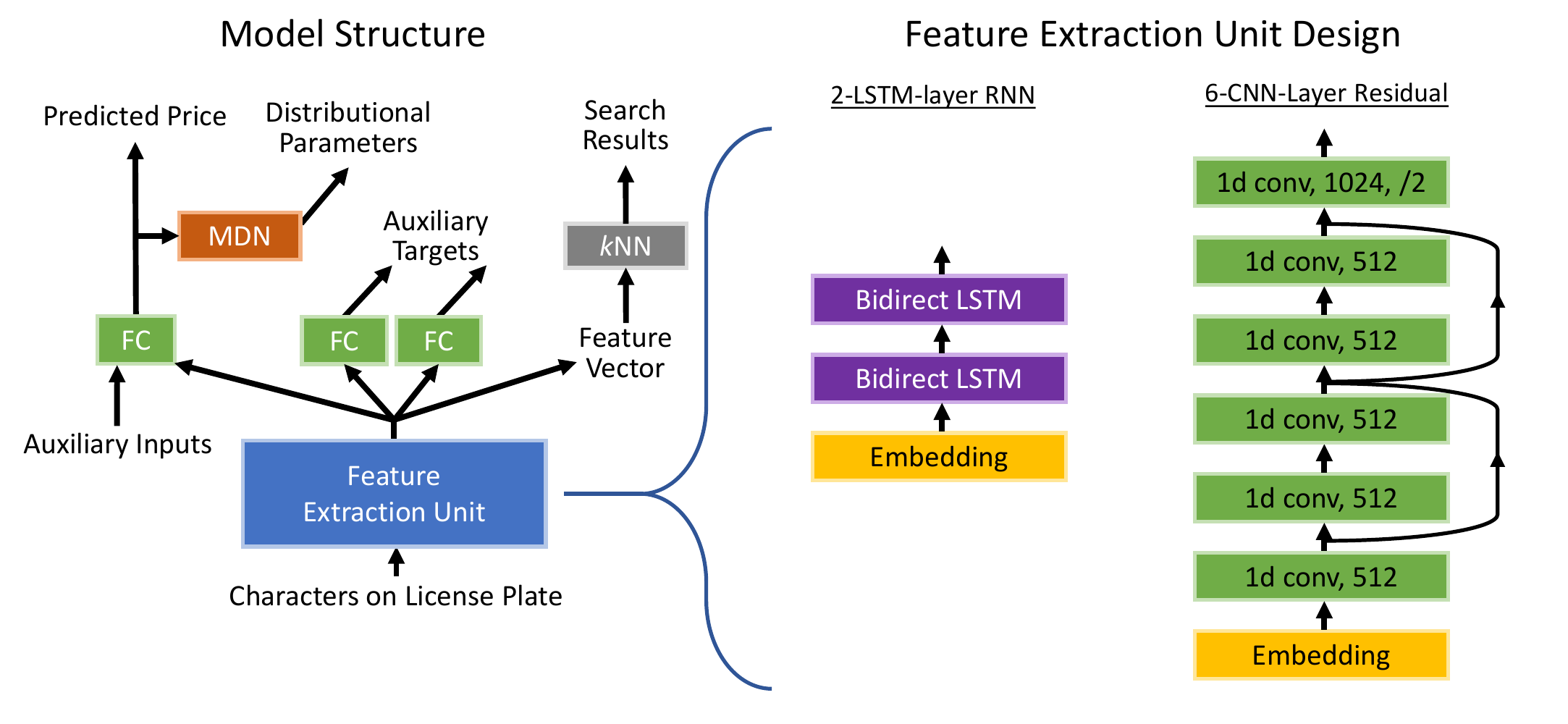}
\caption{Model structure and examples of feature extraction unit design}
\label{wireframe}
\end{figure}

Figure \ref{wireframe} illustrates the structure of our model.
Each plate is represented by a vector of numerically-encoded characters, padded to the same length.
The characters are fed into a feature extraction unit, either one at a time for RNN or as a vector for CNN.
The feature extraction unit outputs a feature vector, which is used in three ways.
First, it is concatenated with auxiliary inputs and fed into a set of fully-connected layers to generate the predicted price. 
Second, it is fed into another set of fully-connected layers to generate a number of  auxiliary targets.
Third, it is fed into a $k$-nearest-neighbor clustering model to produce a list of similar plates. 
Finally, the predicted price is fed into another fully-connected layer responsible for producing distributional parameters
for the final price distribution. 

\subsection{Feature extraction unit}

The feature extraction unit begins with an embedding layer $g(s)$, 
which converts each character $s$ to a vector representation $\vec{h}_s$: 
$
g(s) = \vec{h}_s \equiv [h_{s}^1,...,h_{s}^n].
$
The dimension of the character embedding, $n$, is a hyperparameter. 
The values $h_{s}^1,...,h_{s}^n$ are initialized with random values and learned through training. 
We experiment with $n$ ranging from 8 to 24 as well as replacing learned embedding with one-hot encoding.

The embedding layer is followed by one or more layers of neurons. We explore three architectures:
\begin{enumerate}
\item
\textit{RNN.} As a baseline, we adopt the model used in \cite{license-plate-RNN}, 
which has been shown to perform significantly better than simpler models such as hedonic regressions and n-grams.
Specifically, the unit consists of one or more bi-directional, 
batch-normalized recurrent layers with rectified linear units as activations.
The feature vector is the sum of the last layer's recurrent output.
We conduct a hyperparameter search with the number of layers varying from 1 to 7 and
the number of neurons varying from 128 to 1024.

\item
\textit{LSTM.} It is widely recognized that Long Short-Term Memory (LSTM) networks
performs better than simple recurrent networks in deep learning \cite{LSTM}. 
Following standard practice, we implement the unit as one  or more bi-directional, 
batch-normalized LSTM layers with logistic and tanh activations. 
The feature vector is the output from the last time step of the last LSTM layer.
As with RNN, we conduct a hyperparameter search with the number of layers varying from 1 to 5 and
the number of neurons varying from 128 to 1600.   

\item
\textit{Residual CNN.}
Our implementation is a 1-demensional version of ResNet \cite{ResNet}, 
illustrated in the rightmost panel of figure \ref{wireframe}.
The main features are residuals being added to the output after every two layers,
and the number of filters being doubled whenever there is a 50 percent down sampling. 
All layers are batch normalized and activated by exponential linear units, 
the latter having been shown to improve training speed and accuracy \cite{ELU}. 
We conduct a hyperparameter search with the number of layers varying from 1 to 7
and the number of filters in the first layer varying from 64 to 1024.
\end{enumerate}

\subsection{Auxiliary inputs}

The auxiliary inputs are the date and time of the auction, 
the most-recent general price level index on the day of the auction,
the local stock market index and the return of the index in the past year and the past month.
Historical values are used for training, but our actual product utilizes up-to-date data.

\subsection{Auxiliary targets}
Previous versions of our model did not have auxiliary targets,
but we discovered that while prediction accuracy was similar across different runs of the same model,
the feature vectors generated were not. 
In hindsight, this should be expected: neural network training is non-convex problem,
and there is no reason why the model should settle on a particular set of latent factors after every training run. 
This imposed a significant problem for our search engine, as search results would at times change noticeably
after retraining the model with new data. 

Our solution to these problem is to train the model with auxiliary targets, 
based on the handpicked features of \cite{Ng2010293}.
The idea is to provide guidance on what features users might be looking for,
while still allowing the model enough flexibility to learn additional features.
Table \ref{t-aux-targets} lists the 32 objective measures of plate characteristics, which include
the number of characters,
the number of repeated characters and
whether the combination of characters is symmetric or sequential.

\begin{table}[]
\small
\centering
\caption{Auxiliary targets}
\label{t-aux-targets}
\begin{tabular}{lllllll}
\toprule
\multicolumn{1}{c}{Letters}          & \multicolumn{6}{c}{Numbers} \\
\cmidrule(l{2pt}r{2pt}){1-1}  \cmidrule(l{2pt}r{2pt}){2-7}
Repeated letters &  \textit{x}00       	& abab & aab  & aaa  	& \# of 0's & \# of 5's \\
No letters       & \textit{x}000         & aaab & abb  & aaaa 	& \# of 1's & \# of 6's \\
= HK              & symmetric        & abbb & abcd & aabb 		& \# of 2's & \# of 7's \\
= XX               &  contains ``13''    	& aaba & dcba &   			& \# of 3's &  \# of 8's \\
                 &  = 911		& abaa & aa   &   			& \# of 4's & \# of 9's    \\
\bottomrule
\end{tabular}
\end{table}
 
\subsection{Mixture density network}
A common way of estimating a distribution is to divide the target samples into mutually-exclusive categories and train a classifier. 
That does not work in our case because during inference the target is not bounded from above---there is always the possibility of a record-breaking price.
Instead, we uses a mixture density network (MDN) to generate the distributional parameters \cite{MDN}.

The estimated probability density function of the realized price $p$ for a given predicted price $\hat{p}$ is modelled as a Gaussian mixture:
\begin{equation}
P(p \mid \hat{p}) = \sum_{k=1}^{24}{\frac{e^{z_k(\hat{p})}}{\sum_{i=1}^{24}{e^{z_i(\hat{p})}}} \phi(p \mid \mu_k(\hat{p}),\sigma_k(\hat{p}))},
\end{equation}
where $\phi$ represents the standard normal probability density distribution and
$[z_1(\hat{p}),...,z_{i}(\hat{p}),\mu_1(\hat{p}),...,\mu_{i}(\hat{p}),\sigma_1(\hat{p}),...,\sigma_{i}(\hat{p})]$ 
the output vector from a single fully-connected layer with a single input $\hat{p}$. 
$\sigma$'s have exponential activations to ensure that they take on positive values,
while $\mu$'s and $z$'s have linear activations. We conduct a hyperparameter search with $n$ varying from 3 to 24 
and the number of hidden neurons varying from 64 to 256.

\section{Experiment setup}

\subsection{Data}
The data used in this study come from the Hong Kong government and are an extension 
of the dataset used in \cite{Woo1994389,Woo200835,Ng2010293}.
They cover Hong Kong traditional license plate auctions from January 1997 to February 2017. 
To include as many ultra-expensive plates as possible, we also include the 10 most expensive plates since auctions commenced in 1973, 
information that is publicly available online. 
The data consist of 104,994 auction entries, almost twice that of previous studies. Each entry includes 
i. the characters on the plate, 
ii. the sale price (or a specific symbol if the plate was unsold), and 
iii. the auction date.

The distribution of prices is highly skewed---while the median sale price is \$641, the mean sale price is \$2,064.
The most expensive plate in the data is ``28,'' which sold for \$2.3 million in February 2016. 
Following previous studies, we compensate for this skewness by using log prices within the model.
The use of log price also means that the loss function depends on relative error rather absolute error.

Plates start at a reserve price of at least HK\$1,000 (\$128.2). 
The existence of reserve prices means that not every plate is sold, and 12.6 percent of the plates in our data were unsold. 
Because these plates do not possess a price, we follow previous studies and drop them from the dataset, 
leaving us 91,784 entries. The finalized data are randomly divided into three parts: 64 percent for training, 
16 percent for validation and 20 percent for the final hold-out test. 

\subsection{Training}

Continuous outputs of the model are trained using mean-squared error as the loss function,
while binary outputs are trained using cross entropy as the loss function. 
The primary target and the whole of the auxiliary targets carry equal weight in the overall loss function.
To allow ourselves the flexibility to use a different training duration for the mixture density network, 
the latter is trained separately after other parts of the model have been trained, 
using the negative log-likelihood of the Gaussian mixture as the loss function.

To compensate for the scarcity of expensive plates, we weight samples according to their log price.
We further overweight the most expensive plates, specifically those with a log price above 12.5
(approximately \$34402), by a factor of 40.
These weights are used in all training runs without further experimentation. 

Drop out is applied to each layer in the feature extraction unit except the embedding layer. 
When we started our experiment with RNN we experimented with drop out rate ranging from 0 to 30 percent,
but by the time we reached CNN we have decided to settle on 15 percent since 
there is little variation between different positive drop out rates---excluding a rate of zero,
the correlation between validation RMSE and drop out rate is only 0.003.

In total, we have 432 sets of hyperparameters for both RNN and CNN. 
Due to time constraint, we had to do a relatively sparse search for LSTM with only 95 sets of hyperparameters.
We train the model under each design and each set of hyperparameters three times, 
with early stopping and reloading of the best state.
Recurrent versions of the model are trained for 120 epochs while convolutional versions are trained for 800.
We pick these numbers because trials we conducted before this study 
suggest these values are large enough for the model to almost certainly stop early.
The mixture density network is trained for 5000 epochs with reloading of the best state.

The Adam optimizer with a learning rate of 0.001 is used throughout \cite{adam}. 
Training is conducted with NVIDIA GTX 1080s with mini-batch size of 2,048. 

\section {Experiment results}

\subsection{Predicted price}
\label{model_perf}

\begin{table}
\scriptsize
\centering
\caption{Model performance}
\label{tbl-model_perf}
\begin{tabular}{@{}lllllll@{}}
\toprule
Configuration	&	Train RMSE	&	Valid RMSE	&	Test RMSE  &	Train $R^2$	&	Valid $R^2$	&	Test $R^2$  \\
\midrule
Residual CNN		&	.3859	&	.4692	&	.4721	&	.8985	&	.8499	&	.8471	\\
LSTM	&	.4747	&	.5007	&	.4982	&	.8464	&	.8290	&	.8297	\\
RNN	&	.5069	&	.5443	&	.5492	&	.8473	&	.8197	&	.8237	\\
&&&&&& \\

Woo et al. (2008)	&	.6739	&	.6808	&	.6769	&	..6905	&	.6840	&	.6857 \\
Ng et al. (2010)	&	.6817	&	.6880	&	.6856	&	.6833	&	.6773	&	.6775 \\
unigram $k$NN-10	&	.8924	&	1.174	&	1.165	&	.4572	&	.0599	&	.0690 	\\
\bottomrule

\end{tabular}

\medskip

\begin{minipage}{0.9\textwidth} 
\footnotesize{
For Residual CNN, LSTM and RNN, the average numbers from the best-performing set of hyperparameters are reported.
}
\end{minipage}
\end{table}

Table \ref{tbl-model_perf} lists the performance figures of the best model in each category and that of 
a number of simpler models for comparison.
The best performing model is a 6-layer ResNet, with 512 filters per layer in the first five layers and a 50 percent
down-sampling in the last, paired with a single 256-neuron fully-connected layer and an 8-channel embedding.
This is followed by the bi-directional LSTM with 3200 neurons and two fully-connected layers of 512 neurons each. 
The basic bidirectional RNN comes in last among the three, performing at its best when there is a single recurrent
layer of 1024 neurons and three fully-connected layers of 1024 neurons each. 
Both recurrent models perform best with one-hot encoding.
The best convolutional model is able to explain a significantly higher fraction of variation in prices than the best recurrent models, 
both in sample (5.2 percent as measured by R-squared) and out of sample (1.7 percent).

\begin{figure}
\centering
	\begin{minipage}{0.5\textwidth}
		\centering
		\includegraphics[width=2in]{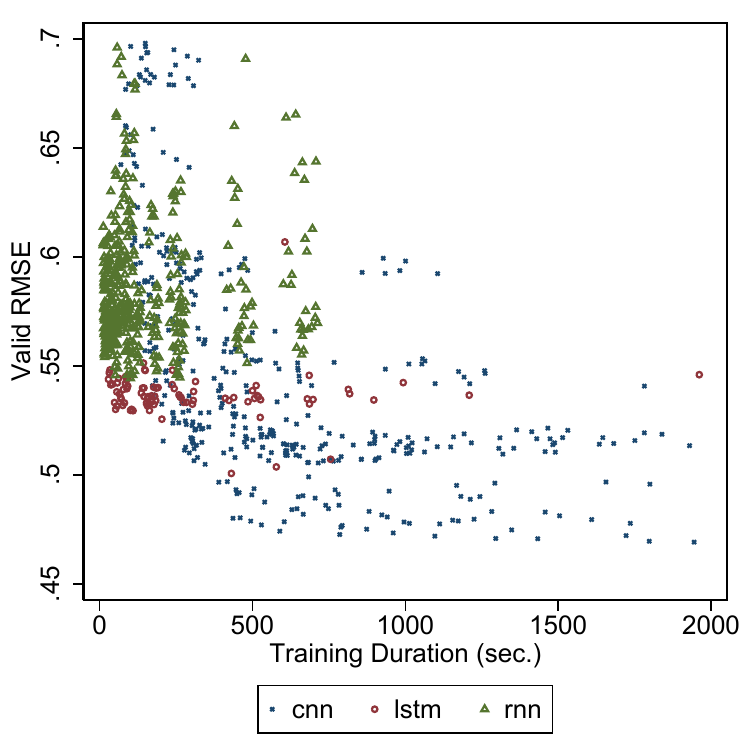}
	\end{minipage}\hfill
	\begin{minipage}{0.5\textwidth}
		\centering
		\includegraphics[width=2in]{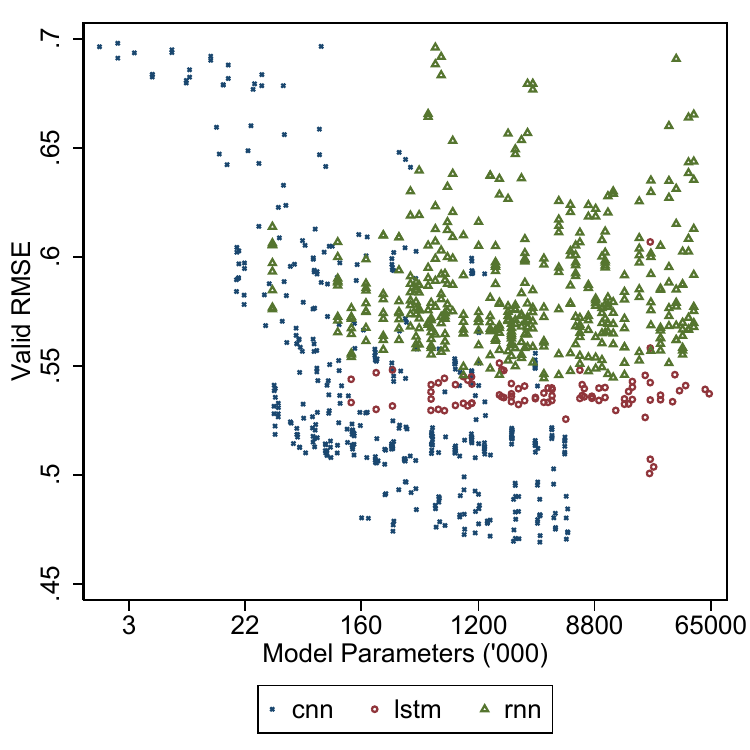}
	\end{minipage}


\caption{Performance as a function of training time and model complexity}
\label{g_structures_compared}

\end{figure}

Figure \ref{g_structures_compared} plots validation RMSE against training duration and complexity of a given model.
Each marker is the average from all runs of a particular set of hyperparameters.
Except for the smallest convolutional networks, 
it is clear from the plots that convolutional networks outperform recurrent networks for either comparable training time or comparable model complexity.
One interesting observation from \ref{g_structures_compared} is that the relationship
between a model's performance and its training time/complexity is much more pronounced for convolutional networks than 
recurrent networks.

To better understand why convolutional networks perform better than recurrent networks, 
Table \ref{tbl-model_perf_pc} lists the error figures for three linguistic patterns and three visual patterns.
Although the best LSTM model is behind the best CNN model in all cases, 
the differences in error are much smaller for linguistic patterns than for visual patterns.

\begin{table}
\scriptsize
\centering
\caption{Model performance by plate characteristics}
\label{tbl-model_perf_pc}
\begin{tabular}{@{}llrrrr@{}}
\toprule
&&	CNN RMSE	&	LSTM RMSE	&	Absolute Diff.	&	Relative Diff.	\\
\midrule
\multicolumn{2}{l}{\textit{Linguistic Patterns}} &&&& \\									
\hspace{0.3cm} 168	& (all the way to prosperity)	&	.3284	&	.4156	&	.0872	&	27\%	\\
 \hspace{0.3cm} 28	& (easy prosperity)	&	.3662	&	.4291	&	.0629	&	17\%	\\
\hspace{0.3cm} 1314	& (together forever)	&	.6049	&	.7000	&	.0951	&	16\%	\\
\multicolumn{2}{l}{\textit{Visual Patterns}} &&&& \\									
\hspace{0.3cm} abba	&&	.4466	&	.6139	&	.1673	&	37\%	\\
\hspace{0.3cm} abcd	&&	.4753	&	.6479	&	.1726	&	36\%	\\
\hspace{0.3cm} aabb	&&	.4934	&	.8006	&	.3071	&	62\%	\\
\bottomrule
\end{tabular}
\end{table}

\begin{figure}
\centering
	\begin{minipage}{0.5\textwidth}
		\centering
		\includegraphics[width=2.3in]{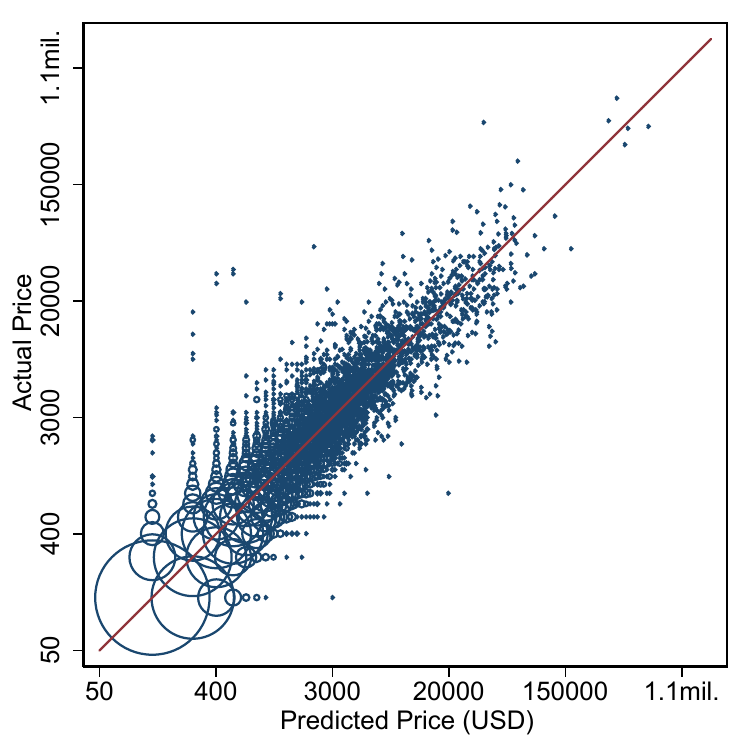}
	\end{minipage}\hfill
	\begin{minipage}{0.5\textwidth}
		\centering
		\includegraphics[width=1.5in]{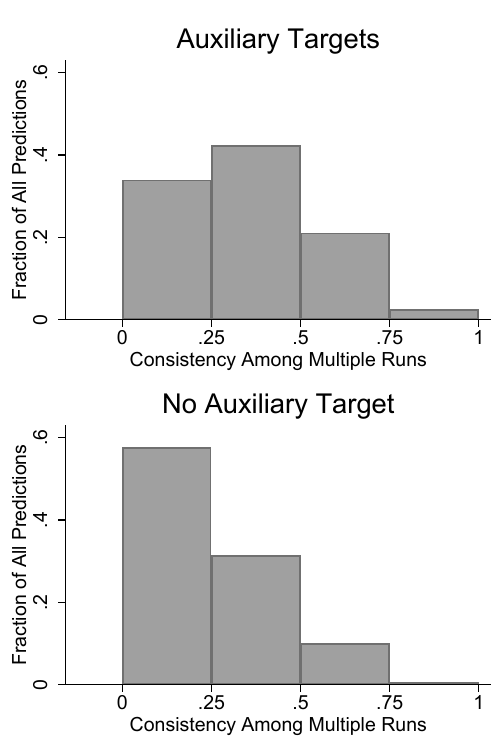}
	\end{minipage}


	\begin{minipage}{0.5\textwidth}
		\caption{Actual vs predicted price in test set}
		\label{actual_v_predicted}
	\end{minipage}\hfill
	\begin{minipage}{0.5\textwidth}
		\caption{Consistency of recommendations}
		\label{g-recom-pvar}
	\end{minipage}

	\medskip

	\centering
	\includegraphics[width=4.8in]{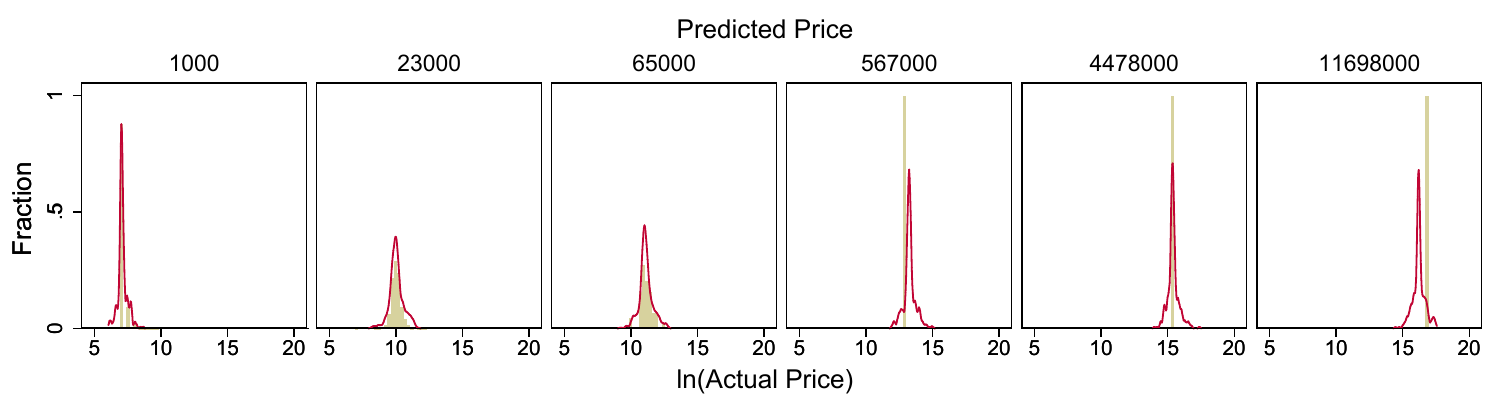}
	\caption{Estimated distribution vs actual distribution at selected predicted prices}
	\label{mdn}

\end{figure}

Figure \ref{actual_v_predicted} plots predicted prices against actual prices from the best model, grouped in bins of HK\$1000 (\$128.2). 
The model performs well for a wide range of prices, with bins tightly clustered along the 45-degree line. 
In particular, due to a better model architecture and weights of samples,
the systemic underestimation of prices for the most expensive plates observed in \cite{license-plate-RNN} is not present here.

\subsection{Feature vector and auxiliary targets}

To demonstrate the effectiveness of the auxiliary targets, 
we rerun the best model with auxiliary targets replaced by the auction price.
This allows us to maintain model complexity while removing the auxiliary targets. 

Table \ref{t-search-consistency} lists the top-three search results for three representative plates
from each of three runs of the best CNN model, with and without the auxiliary targets. 
We measure the consistency of the search results by computing the fraction of search results that appeared in all three runs.
The examples listed in the table illustrate how training the model with auxiliary targets significantly increase consistency across different
 runs of the same model.

\begin{table}[]
\centering
\caption{Auxiliary targets and search consistency}
\label{t-search-consistency}
\scriptsize
\begin{tabular}{lcccccccccc}
\toprule
\textit{}            & \textbf{RMSE} & \multicolumn{9}{c}{\textbf{Search Results}}                                                                  \\ 
\cmidrule(l{2pt}r{2pt}){2-2}  \cmidrule(l{2pt}r{2pt}){3-11}
& & & & & & & & & \\
 \multicolumn{2}{l}{\textbf{With auxiliary targets}}          & \multicolumn{3}{c}{\textit{2112}} & \multicolumn{3}{c}{\textit{BB239}} & \multicolumn{3}{c}{\textit{LZ3360}} \\
 \cmidrule(l{2pt}r{2pt}){1-2}  \cmidrule(l{2pt}r{2pt}){3-5} \cmidrule(l{2pt}r{2pt}){6-8} \cmidrule(l{2pt}r{2pt}){9-11}

\textit{Run 1}       & 0.4750        & 2012      & 1812      & 2121      & CC239      & AA239     & AL239     & HV3360     & BG3360     & HC3360    \\
\textit{Run 2}       & 0.4681        & 1012      & 2012      & 1812      & CC239      & AA239     & LL239     & HV3360     & BG3360     & HC3360    \\
\textit{Run 3}       & 0.4671        & 1812      & 1012      & 2113      & AA239      & CC239     & PP239     & HV3360     & BG3360     & ND6330    \\ 
 \cmidrule(l{2pt}r{2pt}){1-2}   \cmidrule(l{2pt}r{2pt}){3-5} \cmidrule(l{2pt}r{2pt}){6-8} \cmidrule(l{2pt}r{2pt}){9-11}
\textit{Consistency} &               & \multicolumn{3}{c}{0.33}          & \multicolumn{3}{c}{0.67}           & \multicolumn{3}{c}{0.67}            \\ 
& & & & & & & & & \\
 \multicolumn{2}{l}{\textbf{Without auxiliary targets}}           & \multicolumn{3}{c}{\textit{2112}} & \multicolumn{3}{c}{\textit{BB239}} & \multicolumn{3}{c}{\textit{LZ3360}} \\
 \cmidrule(l{2pt}r{2pt}){1-2}  \cmidrule(l{2pt}r{2pt}){3-5} \cmidrule(l{2pt}r{2pt}){6-8} \cmidrule(l{2pt}r{2pt}){9-11}
\textit{Run 1}       & 0.4884        & 9912      & 2223      & 8182      & AA239     & CC199     & AA3298     & HV3360     & HC3360     & JA6602    \\
\textit{Run 2}       & 0.4771        & 1212      & 1812      & 2012      & CC239     & BB989     & AA239      & KE9960     & FE9960     & JR6360    \\
\textit{Run 3}       & 0.4749        & 2832      & 8122      & 8182      & CC239     & AA239     & AA269      & HV3360     & FM6369     & JR6360    \\
 \cmidrule(l{2pt}r{2pt}){1-2}   \cmidrule(l{2pt}r{2pt}){3-5} \cmidrule(l{2pt}r{2pt}){6-8} \cmidrule(l{2pt}r{2pt}){9-11}
\textit{Consistency} &               & \multicolumn{3}{c}{0}             & \multicolumn{3}{c}{0.33}           & \multicolumn{3}{c}{0}               \\ \bottomrule
\end{tabular}
\begin{minipage}{0.93\textwidth}
	\scriptsize{
	Each row is one training run. For each run,
	the top three search results are listed for each plate queried (\textit{2112}, \textit{BB239}, \textit{LZ3360}). 
	Consistency is calculated as the fraction of matches that appear in all three runs.
	}
\end{minipage}
\end{table}

To evaluate consistency systemically, we generate 1000 random new plates and feed them through all six runs of the model.
Figure \ref{g-recom-pvar} plots the consistency measure's distribution with and without the auxiliary targets.
The search results are much more inconsistent when there is no auxiliary target (Mann-Whitney $z=-11.2$, $p=0.0000$), 
with significantly more cases of zero consistency.

\subsection{Estimated Price Distribution}
Figure \ref{mdn} plots the estimated price distribution at selected prices from a mixture density network of 256 hidden neurons 
and a mixture of six Gaussian densities, approximately evenly spaced on a log scale.
The red lines represent the estimated density for a given predicted price, while the bars represent the actual distribution of prices 
for the predicted price. 
The estimated density closely resembles the actual distribution for common, relatively low-value plates.
For very expensive plates, the model is able to produce a density even if there is only a single sample at a given price. 
As can be seen from the examples, the spread of the estimated density generally covers any actual price that deviates from the 
predicted price.

\section{Conclusion}

We demonstrate that a model based on residual convolutional neural net
can generate accurate predicted prices and produce stable feature vectors for use in a search engine.
We demonstrate that for comparable training time or model complexity, a model base on 
convolutional neural nets outperforms one that is base on recurrent neural network.


\bibliographystyle{splncs04}
\bibliography{vinci_ML}

\begin{thebibliography}{10}
\providecommand{\url}[1]{\texttt{#1}}
\providecommand{\urlprefix}{URL }
\providecommand{\doi}[1]{https://doi.org/#1}

\bibitem{10.1257/jep.3.3.23}
Ashenfelter, O.: How auctions work for wine and art. Journal of Economic
  Perspectives  \textbf{3}(3),  23--36 (September 1989).
  \doi{10.1257/jep.3.3.23},
  \url{http://www.aeaweb.org/articles?id=10.1257/jep.3.3.23}

\bibitem{MDN}
Bishop, C.: Mixture density networks. Technical report  (1994)

\bibitem{license-plate-RNN}
Chow, V.: Predicting auction price of vehicle license plate with deep recurrent
  neural network. CoRR  \textbf{abs/1701.08711} (2017),
  \url{http://arxiv.org/abs/1701.08711}

\bibitem{ELU}
Clevert, D., Unterthiner, T., Hochreiter, S.: Fast and accurate deep network
  learning by exponential linear units (elus). CoRR  \textbf{abs/1511.07289}
  (2015), \url{http://arxiv.org/abs/1511.07289}

\bibitem{ResNet}
{He}, K., {Zhang}, X., {Ren}, S., {Sun}, J.: {Deep Residual Learning for Image
  Recognition}. ArXiv e-prints  (Dec 2015)

\bibitem{LSTM}
Hochreiter, S., Schmidhuber, J.: Long short-term memory  \textbf{9},  1735--80
  (12 1997)

\bibitem{adam}
Kingma, D.P., Ba, J.: Adam: {A} method for stochastic optimization. CoRR
  \textbf{abs/1412.6980} (2014), \url{http://arxiv.org/abs/1412.6980}

\bibitem{10.2307/1911865}
Milgrom, P.R., Weber, R.J.: A theory of auctions and competitive bidding.
  Econometrica  \textbf{50}(5),  1089--1122 (1982),
  \url{http://www.jstor.org/stable/1911865}

\bibitem{Ng2010293}
Ng, T., Chong, T., Du, X.: The value of superstitions. Journal of Economic
  Psychology  \textbf{31}(3),  293 -- 309 (2010).
  \doi{http://dx.doi.org/10.1016/j.joep.2009.12.002},
  \url{http://www.sciencedirect.com/science/article/pii/S0167487009001275}

\bibitem{Woo200835}
Woo, C.K., Horowitz, I., Luk, S., Lai, A.: Willingness to pay and nuanced
  cultural cues: Evidence from hong kong's license-plate auction market.
  Journal of Economic Psychology  \textbf{29}(1),  35 -- 53 (2008).
  \doi{http://dx.doi.org/10.1016/j.joep.2007.03.002},
  \url{http://www.sciencedirect.com/science/article/pii/S016748700700027X}

\bibitem{Woo1994389}
Woo, C.K., Kwok, R.H.: Vanity, superstition and auction price. Economics
  Letters  \textbf{44}(4),  389 -- 395 (1994).
  \doi{http://dx.doi.org/10.1016/0165-1765(94)90109-0},
  \url{http://www.sciencedirect.com/science/article/pii/0165176594901090}

\end{thebibliography}

\end{document}